# From Visual Perception to Deep Empathy: An Automated Assessment Framework for House-Tree-Person Drawings Using Multimodal LLMs and Multi-Agent Collaboration

Shuide Wen[*1], Yu Sun[2], Beier Ku[3], Zhi Gao[4], Lijun Ma[5],
Yang Yang[†6], and Can Jiao[†7,8]

[1]Shenzhen International Graduate School, Tsinghua University, Shenzhen, China
[2]School of Psychology, Shenzhen University, Shenzhen, China
[3]Jesus College, University of Oxford, Oxford, UK
[4]Shenzhen Institute of Education Sciences, Shenzhen, China
[5]Department of Psychology, School of Public Health and Management,
Guangzhou University of Chinese Medicine, Guangzhou, China
[6]Harbin Institute of Technology, Harbin, China
[7]School of Government, Shenzhen University, Shenzhen, China
[8]The Shenzhen Humanities & Social Sciences Key Research Bases of the Center
for Mental Health, Shenzhen University, Shenzhen, China

wenshuide@sz.tsinghua.edu.cn, 2410161009@mails.szu.edu.cn,
beier.ku@jesus.ox.ac.uk,
career@sz.edu.cn, malj@gzucm.edu.cn, yfield@hit.edu.cn, jiaocan@szu.edu.cn

## Abstract

**Background** The House–Tree–Person (HTP) drawing test, introduced by John Buck in 1948 (Buck, 1948), remains one of the most widely used projective techniques in clinical psychology. Yet it has long faced challenges such as heterogeneous scoring standards, reliance on examiners' subjective experience and a lack of a unified quantitative coding system (Guo et al., 2023).

**Objective** This study seeks to verify the effectiveness of representative multimodal large language models (MLLMs) in recognising and interpreting HTP features, and to develop an automated assessment framework based on multi-agent collaboration to address hallucinations and empathy deficits associated with single models.

**Methods** The research comprised two stages. In Study 1, 307 anonymised HTP drawings previously interpreted by human experts (e.g., Yan Hu, Wang Long) were re-interpreted by Qwen-VL-Plus, and cosine similarity between model-generated and expert interpretations was computed using Doubao-embedding-vision. In Study 2 a multi-agent system was built, incorporating roles such as an *Observer*, *Interpreter*, *Zeitgeist Observer* and *Listener* (Park et al., 2023), to deliver deep assessments of typical cases.

**Results** Quantitative experiments showed that the mean semantic similarity between MLLM interpretations and human expert interpretations was around 0.75 (SD ≈ 0.05); in structurally oriented expert data sets this rose to 0.85, indicating expert-level baseline comprehension. Qualitative analyses demonstrated that the multi-agent system, by integrating social-psychological perspectives and destigmatising narratives, effectively corrected visual hallucinations and produced psychological reports with high ecological validity and internal coherence.

**Conclusions** The findings confirm the potential of multimodal large models as standardised tools for projective assessment. The multi-agent framework proposed here, by dividing roles, decouples "feature recognition" from "psychological inference" and offers a new paradigm for digital mental-health services.

## 1 Introduction

The House–Tree–Person (HTP) drawing test is a projective technique in which individuals are asked to draw a house, a tree and a person, and the resulting images are analysed to uncover unconscious emotions and personality traits. Since its introduction by John Buck in 1948 (Buck, 1948) and subsequent systematisation by Emanuel Hammer in 1958 (Hammer, 1958), the test has been widely used in clinical evaluation, personnel selection and educational counselling. HTP belongs to non-structured projective methods; its scoring and interpretive depth depend heavily on the examiner's personal experience, leading to inconsistent readings of the same graphical symbols (e.g., chimneys, ornaments). A systematic review noted that current studies lack uniform methods for selecting and interpreting drawing indicators (Guo et al., 2023). Although 39 features are significant predictors of mental disorders, their categories, measurement dimensions and interpretive frameworks remain muddled (Guo et al., 2023); for example, a chimney has been interpreted as a sign of family conflict in some studies and of warmth and external support in others (Guo et al., 2023). Such heterogeneity limits test reliability and restricts large-scale application while complicating clinical training and replication.

In recent years, advances in artificial intelligence—especially multimodal large language models (MLLMs) capable of visual understanding and complex reasoning—have made automated psychological assessment possible. However, existing work largely focuses on simple image classification or object detection tasks and lacks deep reasoning about psychodynamic mechanisms; in addition, single models are prone to hallucination and exhibit limited empathy. To address these issues, the present study employs Qwen-VL-Plus together with Doubao-embedding-vision to verify semantic alignment on HTP drawings, and then designs a multi-agent collaboration framework that integrates developmental, personality and social-psychological perspectives to achieve a full pipeline from objective feature recognition to deep psychological understanding.

The paper addresses two core questions:

1. **Consistency validation** — Can state-of-the-art MLLMs "understand" HTP drawings? How closely do their interpretations align with those of human experts?

2. **System construction** — How can a multi-agent architecture combining clinical and social-psychological perspectives generate interpretable psychological reports with clinical value?

## 2 Study 1: Baseline Alignment Between MLLMs and Human Experts

### 2.1 Data sources and preprocessing

We collected 307 HTP drawings and their corresponding expert interpretations. Experts included Yan Hu, Wang Long (Hu, 2019; Wang, 2014), Min Baoquan, Zhang Tongyan and other senior analysts from different schools, including structural and imagistic approaches, ensuring broad representation. All drawings and text were anonymised and formatted for model input.

### 2.2 Experimental tools

1. **Generator: Qwen-VL-Plus** (Bai et al., 2023) – a representative multimodal large language model with enhanced visual coding, spatial perception and multimodal reasoning.

2. **Evaluator: Doubao-embedding-vision** (ByteDance Doubao Team, 2025) – a model that supports bilingual multimodal embedding, producing 2048-dimensional float vectors for both Chinese and English texts.

### 2.3 Experimental procedure

1. **Unified prompting** – All drawings were input into Qwen-VL-Plus using a standardised prompt requiring a psychological interpretation.
2. **Vectorisation** – Both model-generated and expert textual interpretations were embedded via Doubao-embedding-vision to obtain 2048-dimensional vectors.
3. **Similarity computation** – Cosine similarity between the two vectors was calculated (Radford et al., 2021) as the primary measure of AI–expert alignment.

## 2.4 Results

### 2.4.1 Overall consistency

Across the 307 samples the mean cosine similarity between AI and expert interpretations was 0.748 with a standard deviation of 0.058. Over 82% of samples achieved similarity scores above 0.70, indicating that the model captured the core features and psychological themes emphasised by human experts. Figure 1A displays the distribution of case similarities.

### 2.4.2 Mapping expert heterogeneity

Analysis revealed significant differences across expert datasets. For example, the structuralist expert Wang Long produced an average similarity of 0.787 over approximately 178 cases, with low variance; multiple cases exceeded 0.85. This suggests that his interpretations are logically structured and feature associations are clear, making them easier for the model to learn. In contrast, data from imagistic experts such as Yan Hu and Zhang Tongyan yielded mean similarities around 0.65 and included cases as low as 0.556, perhaps because their interpretations rely on cultural imagery and clinical intuition, which are less explicit.

### 2.4.3 Similarity distribution and expert differences

To visualise AI–expert alignment, we plotted four graphics:

- **Figure 1A: Expert mean similarity bar chart** compares mean similarities across experts (Wang Long, Micro Psychological, Yan Hu), highlighting how model performance varies by interpretive style.
- **Figure 1B: Similarity distribution histogram** depicts the distribution of similarity scores across 307 cases, showing the majority concentrate around 0.70–0.80.
- **Figure 1C: Expert group box plot** compares similarity distributions by expert; structuralist data (Wang Long) exhibit higher medians and smaller dispersion.
- **Figure 1D: Expert heterogeneity violin plot** displays violin plots for each expert dataset, illustrating density distributions, medians and extremes. This reinforces the finding that structuralist interpretations are the easiest for the model to learn.

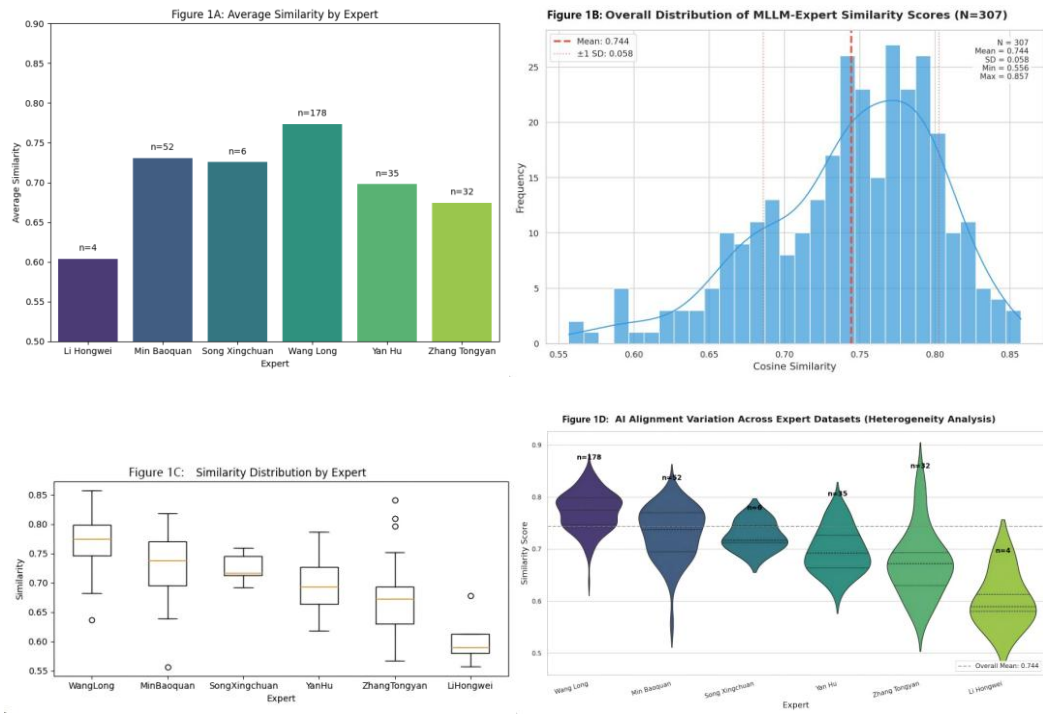

Figure 1: Four subfigures showing AI-expert alignment visualizations

Taken together, Figures 1A–1D show that the vast majority of cases achieve similarity scores ≥ 0.70 and that Wang Long's data yields the highest distribution. Despite differences across experts, MLLMs demonstrate sufficient alignment for practical use.

### 2.4.4 Overall statistics and expert comparison

To present the experimental data more intuitively, we calculated overall statistics for all 307 cases and grouped statistics by expert (Table 1). Metrics include case counts, mean similarity, quartiles (Q1, Q3), medians, maxima, minima and standard deviations. Overall, the mean similarity was 0.7441 with a standard deviation of 0.0583; Q1, median and Q3 were 0.7089, 0.7521 and 0.7882 respectively, indicating that most cases fall between 0.70 and 0.80.

| Expert | Cases | Mean | Median | Q1 | Q3 | Min | Max | SD |
| --- | --- | --- | --- | --- | --- | --- | --- | --- |
| Wang Long | 178 | 0.7735 | 0.7754 | 0.7469 | 0.7991 | 0.6367 | 0.8574 | 0.0369 |
| Min Baoquan | 52 | 0.7310 | 0.7376 | 0.6953 | 0.7702 | 0.5566 | 0.8179 | 0.0498 |
| Song Xingchuan | 6 | 0.7257 | 0.7175 | 0.7131 | 0.7455 | 0.6922 | 0.7602 | 0.0264 |
| Yan Hu | 35 | 0.6978 | 0.6929 | 0.6643 | 0.7272 | 0.6182 | 0.7870 | 0.0425 |
| Zhang Tongyan | 32 | 0.6744 | 0.6722 | 0.6299 | 0.6936 | 0.5678 | 0.8409 | 0.0641 |
| Li Hongwei | 4 | 0.6040 | 0.5899 | 0.5807 | 0.6132 | 0.5579 | 0.6781 | 0.0517 |

Table 1: Summary statistics by expert and overall.

## 3 Study 2: Deep Interpretation via Multi-Agent Collaboration

Study 1 exposed two key challenges: (1) MLLM performance is sensitive to expert style, resulting in heterogeneity; (2) single models lack deep psychodynamic reasoning and social-context understanding. Accordingly, Study 2 develops a multi-agent system to deliver a pipeline from visual feature extraction to comprehensive psychological interpretation.

### 3.1 System architecture

The experimental tools and test platform in this study are built on our team's proprietary **Qingliu Agent** multi-agent collaboration system (Wu et al., 2023). This platform provides specialised functionality—including multi-agent definitions, workflow orchestration, context management, task parallel processing, and high-dimensional vector embedding and clustering—that makes it suitable both for sociological simulation experiments and for engineering deployment.

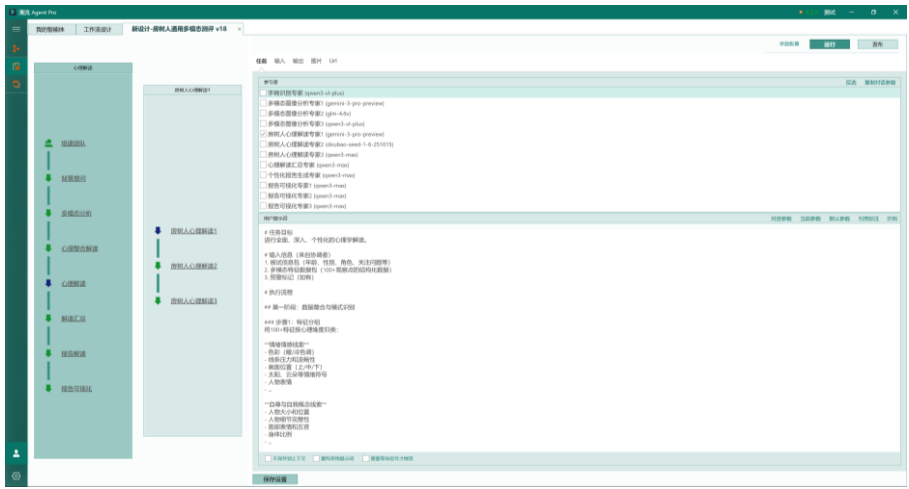

Figure 2: Qingliu Multi-agent System

The system comprises the following roles:

1. **Observer** – objectively detects visual features in the drawing, such as line quality, spatial layout, proportional relationships and elemental details, producing a structured JSON record of more than 150 observation points.

2. **Interpreter** – receives the observer's JSON output and maps visual features to psychological meanings based on Buck/Hammer theory (Buck, 1948; Hammer, 1958) and developmental psychology, building evidential chains and identifying defence mechanisms and personality tendencies.

3. **Zeitgeist Observer** – introduces a social-psychological perspective, situating individual issues within macro social and cultural contexts. It draws on research and statistics to contextualise phenomena such as mid-life crises and professional burnout, reducing pathologisation.

4. **Listener** – integrates information from all modules into a warm, comprehensible and inspiring personalised report, applying positive psychology (Seligman, 2011) to propose growth suggestions and action plans.

Additional agents include a visualisation specialist and report designer, who create interactive graphics and user-friendly presentations. Agents share intermediate data and engage in critical feedback, forming a virtual "expert consultation" to ensure logical coherence and ecological validity.

### 3.2 Case study

We tested the system on an HTP drawing by a 38-year-old male IT worker (ID HTR-38-M-20240520). Whereas a single model produced only generic descriptions, the multi-agent system provided nuanced insights:

- **Feature identification** – The observer accurately extracted details such as "a structurally complete house," "a monochrome winter tree" and "a stick-figure person," and quantified line thickness, positions and paper utilisation.

- **Psychological inference** – The interpreter mapped features to meanings, noting good emotion regulation and a sense of responsibility but also low self-efficacy and depleted life energy, and proposed age-appropriate growth strategies.

- **Societal context** – The Zeitgeist observer cited reports like the *2025 Chinese Workplace Wealth Report* (Zhaopin & DT Finance, 2025) and *Education Anxiety White Paper* to contextualise anxieties about "financial freedom" and "child-rearing"; these are common among post-1985 male IT workers and not necessarily pathological. It offered collective wisdom strategies such as micro-entrepreneurship and side-hustle validation.

- **Empathic report** – The listener composed a warm narrative that affirmed strengths (emotion management, responsibility) before suggesting actionable steps (small success accumulation, sensory awakening exercises, emotional demonstration practice) and short-, medium- and long-term goals. It included suggestions on social support networks and cited social statistics to help the examinee find resonance and resources amid structural challenges.

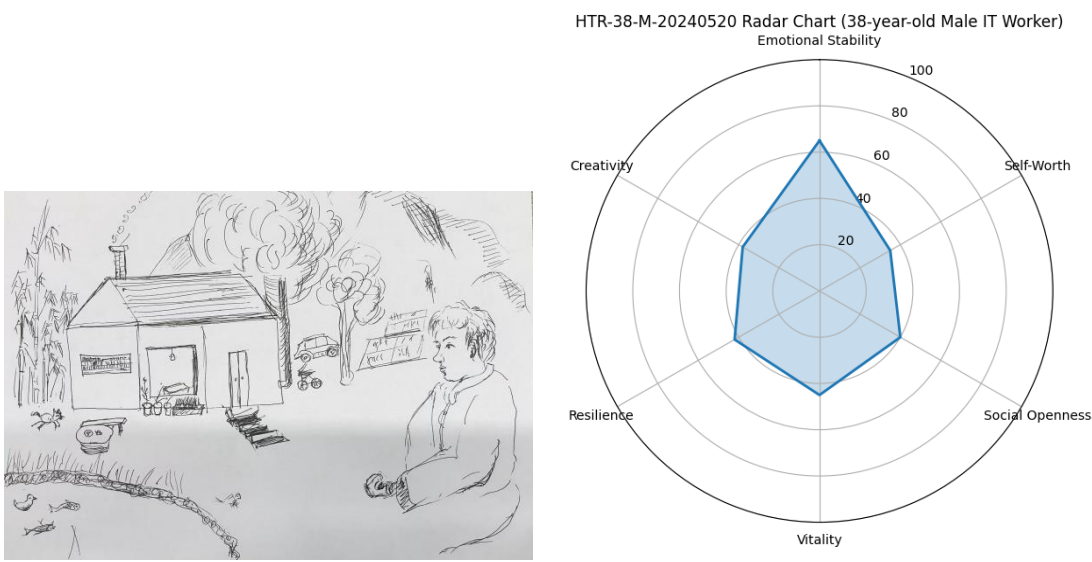

Figure 3: HTR-38-M-20240520 Radar Chart (38-year-old Male IT Worker)

Quantitatively, this case's six-dimensional radar plot showed scores of 65 for emotional stability, 35 for self-worth, 45 for vitality, 42 for resilience and 38 for creativity. The report further addressed the examinee's concerns about "financial freedom," "raising the next generation" and "whether to start a business."

### 3.3 Multi-case analysis

To further validate the applicability of the multi-agent framework, we evaluated over 30 real-world HTP drawing samples and selected two representative participants with distinct backgrounds to generate in-depth psychological reports. Below we outline the observer's objective feature extraction, the interpreter's psychological reasoning and the integrated insights from other agents.

**Case A: 25-year-old female doctoral student**

**Demographics** A 25-year-old woman pursuing a PhD. Her radar chart (Figure 2A)

shows scores of 68 (emotional stability), 52 (self-worth), 60 (social openness), 70 (vitality), 65 (resilience) and 62 (creativity).

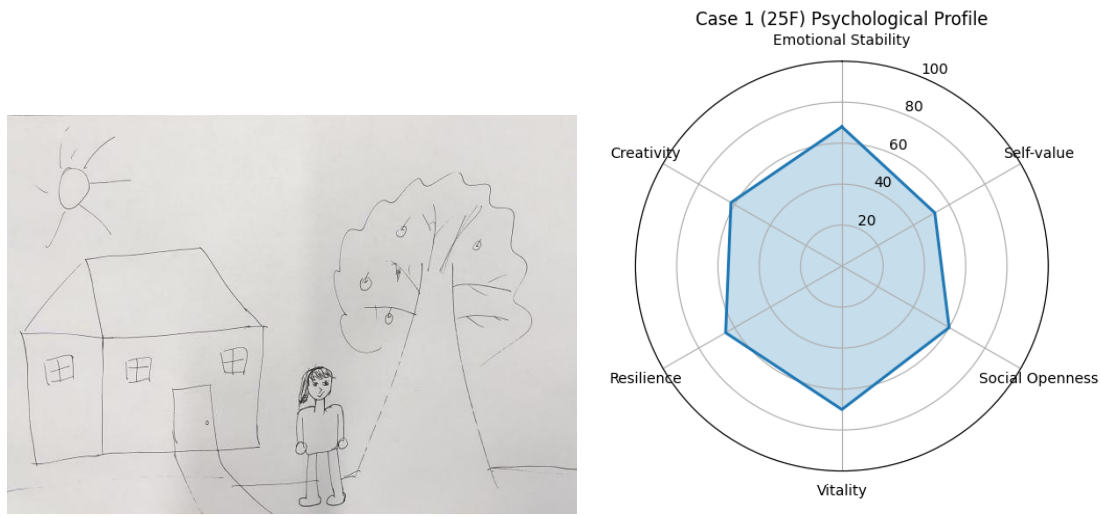

Figure 4: Case 1 (25F) Psychological Profile

**Observer findings** The participant drew a structurally complete house, a fruit-laden tree and a relatively small person. The scene features a sun and a path extending forward; composition is balanced and orderly with smooth lines. The person has round hands without fingers, a simplified neck and a block-like body.

**Interpreter insights** According to the Buck/Hammer framework:

- **High goal orientation and achievement awareness** (vitality = 70). Fruits on the tree and the extending path symbolise a strong drive for achievement and clear goals, providing both academic motivation and psychological resilience.

- **Solid reality base and structured thinking** (resilience = 65; emotional stability = 68). A horizon running through the picture, and stable house and tree trunks, reflect a firm psychological foundation and logical thinking, enabling the person to maintain order under pressure.

- **Rational optimism and hope** (emotional stability = 68). The sun, smiling figure and ripe fruits indicate that despite academic stress she retains hope and a positive outlook.

At the same time, the report highlights growth areas:

- **Transformation of knowledge into action** (self-worth = 52). The simplified round hands suggest an inability to "grasp," implying that although she knows what to do she struggles to start; this manifests as procrastination and perfectionism.

- **Mind–body integration and emotional awareness**. The rigid block body and absent neck indicate a tendency toward intellectualisation and neglect of bodily signals and emotional experience.

**Integrated suggestions** The listener proposed interventions including a "micro-action starter method" (writing 200-word drafts daily), body-scan mindfulness exercises, recording daily signs of hope and building support networks (mentors, peers, family). The Zeitgeist observer, citing 2024–2025 surveys of PhD candidates (iResearch, 2024), noted that most doctoral students face action-initiation difficulties and tie their self-worth to publication outcomes, so these suggestions have broad applicability.

**Case B: 45-year-old male mental-health practitioner**

**Demographics** A 45-year-old man engaged in clinical and educational work, mid-career. His radar chart (Figure 2B) scores 82 (emotional stability), 78 (self-worth), 75 (social openness), 85 (vitality), 80 (resilience) and 76 (creativity).

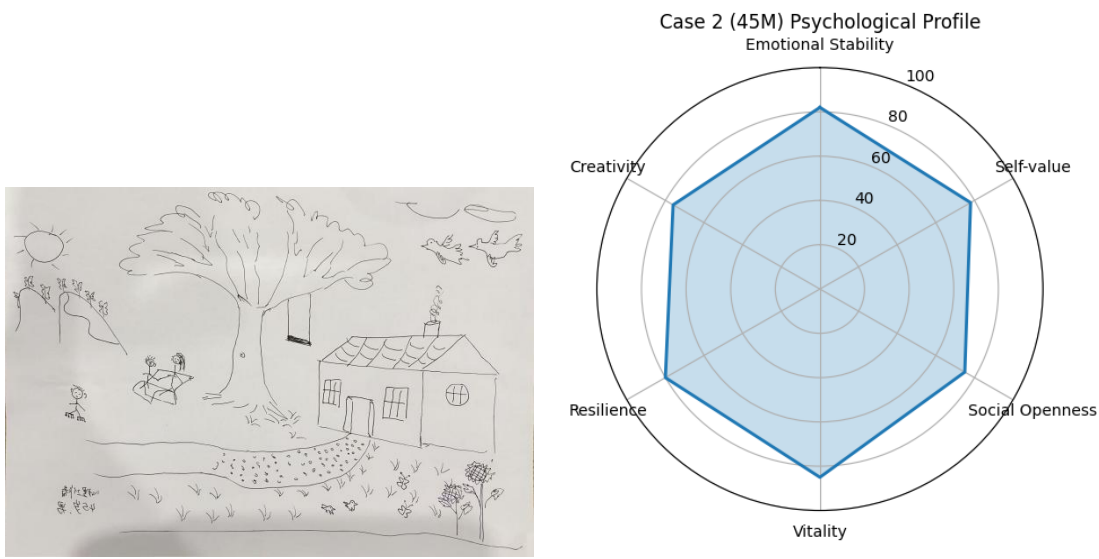

Figure 5: Case 2 (45M) Psychological Profile

**Observer findings** The drawing depicts a tree with a cloud-like crown, a bench and swing beneath it; the house is drawn in perspective with smoke rising from its chimney; a stone path winds into the distance. A person stands to the side wearing roller shoes, rendered as a stick figure. There are no signs of damage or scribbling.

**Interpreter insights**

- **Outstanding system integration and planning ability** (creativity = 76). The

> S-shaped stone path and three-dimensional house indicate that he can convert abstract ideas into systematic plans, balancing grand vision and detailed execution.

- **Strong family responsibility and protective energy** (vitality = 85; emotional stability = 82). The tree shelters the family; the house is stable; smoke from the chimney and smiling figures symbolize providing a safe harbour.

- **Mature defence mechanisms and psychological resilience** (resilience = 80). Scars on the trunk and the swing suggest the ability to integrate trauma and use intellectualisation and symbolic regression to maintain function under stress.

Growth areas include:

- **Limited emotional involvement and authentic self-expression**. The stick-figure body and separate stance imply that he often takes the role of an observer, offering support but seldom revealing vulnerability, potentially impacting intimacy.

- **Balancing productive roles and self-care**. Roller shoes indicate constant readiness to rush outward; the swing reflects a desire for rest, revealing tension between work and self-nurture.

**Integrated suggestions** The system advised scheduling non-goal-oriented parent–child time and institutionalising "swing time"; building teacher mutual-aid groups and ecological teaching-research systems to share responsibilities; and incorporating narrative perspectives in research to avoid over-intellectualisation. The Zeitgeist observer noted that 45-year-old mental-health professionals face dual pressures of career transition and family responsibility; these strategies address the "sandwich generation" anxiety.

Through these cases, the multi-agent system demonstrates adaptability and empathy across ages and professions. From feature extraction and psychological inference to social context inclusion and empathic reporting, the system offers personalised, developmental guidance, showcasing the potential of AI-assisted psychological assessment.

### 3.4 Clinical validity and advantages

Comparing outputs from single models and the multi-agent system reveals that the latter excels in logical coherence, empathy quality and cultural adaptation. The multi-agent framework effectively combines expertise from different disciplines, reducing hallucinations and generating structured data and personalised interpretations. In clinical practice, the system can be used not only for assessment but also to provide targeted intervention suggestions and resource links, expanding the utility of the HTP test.

**3.5 Multimodal model fusion and integrated report synthesis**

Although multi-agent collaboration enriches interpretations beyond a single model, any individual visual-language model may still harbour bias. To mitigate this, Study 2 also employed parallel model interpretation and text amalgamation. We selected three representative multimodal models—Gemini-3-Pro-Preview, Qwen3-VL-Plus and Doubao-Seed-1-6-Vision—to independently interpret each HTP drawing; their outputs were generated by three "multimodal image analysis experts."

We then used DeepSeek-v3.1-Terminus (DeepSeek-AI, 2024), a powerful text model, to merge the three reports. The prompt required DeepSeek to follow four principles: **Directness** (do not downplay any psychological concerns or risk signals), **Evidence** (all conclusions must be grounded in specific observations and cite their sources), **Caution** (distinguish between highly consistent, partially supported and divergent views) and **Practicality** (provide operational directions for subsequent support or intervention).

The merging process comprised four steps: (1) extract core viewpoints from the three interpretations across six dimensions—house, tree, person, overall layout, stroke features and special symbols/omissions; (2) identify consensus and divergence, classifying consensus as reliable findings and unique insights as "to be verified," and analysing conflicting views; (3) conduct risk assessments focusing on emotional distress, self-perception, interpersonal difficulties, developmental concerns and red flags requiring professional attention; (4) generate an integrated report using a predefined format, including an executive summary, detailed analysis, priority concerns and warnings, support recommendations and limitations.

This multi-model fusion harnesses complementary strengths across models, reducing random errors and enhancing robustness and clinical applicability. For instance, in the 45-year-old practitioner case, the synthesis found that the "complete house with smoke" symbolising safety was consistently noted by all models, whereas a "scar on the trunk" appeared in only one interpretation and was flagged as a unique observation requiring verification. The final report agreed that stick-figure representation and the swing indicated emotional avoidance and regression and provided immediate and long-term support suggestions. Through this workflow the system can produce more comprehensive, objective and clinically useful interpretations, offering a new paradigm for digital psychological assessment.

## 4 Discussion

### 4.1 AI standardisation and heterogeneity

Study 1 showed that the mean cosine similarity between MLLM interpretations and expert interpretations is about 0.75 and is widely distributed above 0.7. This suggests that AI can serve as an anchor for HTP scoring. Standardised coding by the observer agent unifies feature recognition across experts and reduces human bias. However, expert styles vary; single models struggle to adapt to all schools. The multi-agent system alleviates this by dividing roles and building evidential chains. Moreover, our framework introduces a multimodal fusion mechanism (Section 3.5) that cross-validates outputs from three different MLLMs via DeepSeek, further enhancing reliability.

Systematic reviews have identified 50 repeatedly reported drawing features, with 39 significantly associated with mental disorders and the "overall drawing" dimension having the strongest predictive effect (odds ratio ≈ 4.20) (Guo et al., 2023). Yet researchers lack a unified standard for selecting and interpreting these features, and even disagree on the meaning of the same symbol (Guo et al., 2023). Our multi-agent framework, through standardised feature extraction and evidential reasoning, aims to harmonise indicator systems and reduce such heterogeneity.

### 4.2 Emergent capabilities and empathy

Multi-agent collaboration exhibits emergent abilities not present in single models: the observer ensures objectivity, the interpreter provides psychological inference, the Zeitgeist observer introduces social context and the listener crafts an empathic narrative. This layered approach allows the system not only to identify objective features but also to understand individuals' situations and offer developmental guidance. By incorporating social statistics (e.g., collective anxieties about wealth and child-rearing among post-1985 IT workers), the system helps examinees shift from "I am flawed" to "I am not alone," which traditional assessments seldom achieve.

### 4.3 Limitations and future directions

Several limitations remain. Qwen-VL-Plus performs poorly on extremely abstract or highly artistic drawings; specific domain adaptation or fine-tuning is needed. Our dataset focuses on adult participants in China; future work should expand across ages and cultures to test generalisability. Furthermore, the multi-agent system's decision process is complex; more efficient scheduling strategies and model-fusion methods are needed.

## 5 Conclusions

We developed a House–Tree–Person assessment framework based on multimodal large models and multi-agent collaboration. By analysing 307 expert interpretations, we demonstrated baseline comprehension by MLLMs (similarity ≈ 0.75) and uncovered expert-style heterogeneity. Building on this, the multi-agent system delivers a complete pipeline from objective feature recognition to deep psychological empathy, introducing social-psychological perspectives for more humanised reports. This framework offers a feasible path to standardising and digitising projective tests and broadens the application of AI in mental-health assessment.